\documentclass[lettersize,journal,hidelinks,bookmarksopen=true,pdfstartview=FitH]{IEEEtran}
\usepackage{amsmath,amsfonts}
\usepackage{algorithmic}
\usepackage{array}
\usepackage[caption=false,font=footnotesize]{subfig}
\usepackage{textcomp}
\usepackage{stfloats}
\usepackage{url}
\usepackage{verbatim}
\usepackage{graphicx}
\usepackage{bm}
\usepackage{balance}
\usepackage{orcidlink}
\usepackage{cite}
\usepackage{tikz}
\usepackage{multirow}
\usepackage{xspace}
\usepackage{diagbox}

\def\BibTeX{{\rm B\kern-.05em{\sc i\kern-.025em b}\kern-.08em
		T\kern-.1667em\lower.7ex\hbox{E}\kern-.125emX}}

\graphicspath{ {./} }

\newcommand*{\eg}{e.g.\@\xspace}
\newcommand*{\ie}{i.e.\@\xspace}
\newcommand*{\etal}{et al.\@\xspace}
\newcommand*{\deltat}{$\Delta t$}
\newcommand*{\rsc}{$r_\mathrm{s,c}$\@\xspace}
\newcommand*{\rsq}{$r_\mathrm{s,q}$\@\xspace}
\newcommand*{\rscell}{$r_\mathrm{s,cell}$\@\xspace}
\newcommand*{\lorawan}{LoRaWAN\@\xspace}
\newcommand*{\isquareds}{I\textsuperscript{2}S\@\xspace}

\begin{document}
	\title{\lorawan Based Dynamic Noise Mapping with Machine Learning for Urban Noise Enforcement}
	\author{
		\IEEEauthorblockN{
			H.~Emre~Erdem%
			,~Henry~Leung
		}%
		\thanks{This~work~was~supported~in~part~by~the~Alberta~Jobs,~Economy~and~Innovation~Ministry’s~Major~Innovation~Fund~to~the~Centre~for~Autonomous~Systems~in~Strengthening~Future~Communities~and~in~part~by~the~Natural~Sciences~and~Engineering~Research~Council~of~Canada.}%
	}
	
	\markboth{}%
	{How to Use the IEEEtran \LaTeX \ Templates}
	
	\maketitle
			
	\begin{abstract}
		Static noise maps depicting long-term noise levels over wide areas are valuable urban planning assets for municipalities in decreasing noise exposure of residents.
		However, non-traffic noise sources with transient behavior, which people complain frequently, are usually ignored by static maps.
		We propose here a dynamic noise mapping approach using the data collected via low-power wide-area network (LPWAN, specifically LoRaWAN) based internet of things (IoT) infrastructure, which is one of the most common communication backbones for smart cities.
		Noise mapping based on LPWAN is challenging due to the low data rates of these protocols.
		The proposed dynamic noise mapping approach diminishes the negative implications of data rate limitations using machine learning (ML) for event and location prediction of non-traffic sources based on the scarce data.
		The strength of these models lies in their consideration of the spatial variance in acoustic behavior caused by the buildings in urban settings.
		The effectiveness of the proposed method and the accuracy of the resulting dynamic maps are evaluated in field tests.
		The results show that the proposed system can decrease the map error caused by non-traffic sources up to 51\% and can stay effective under significant packet losses.
	\end{abstract}
	
	
	\section{Introduction}
	\label{sec:intro}
	\IEEEPARstart{N}{oise} exposure is long proven to be a primary health problem due to its negative impacts on main body functions such as sleep, hearing, and mental capacity \cite{passchier2000noise}.
	Thus, diminishing noise exposure plays a key role in improved quality of life.
	For this purpose, municipalities around the globe utilize noise management practices to constrain the exposure within healthy limits \cite{jin2014information}.
	These practices include passive measures such as consideration of static traffic noise maps for new development plans as well as active noise enforcement measures such as elimination of noise limit violations.

	\IEEEpubidadjcol

	Static (or strategic) noise maps used by passive measures depict noise levels of desired areas over extended periods (\eg a year) \cite{baclet2022strategic}.
	These levels are calculated based on the methodological frameworks such as CNOSSOS-EU, CRTN, and sonROAD18 created by various government authorities \cite{kephalopoulos2012common, CRTN, heutschi2018sonroad18}.
	Since static noise maps present noise levels over extended periods, their outputs represent only perpetual noise sources such as road traffic and industrial facilities and not the intermittent sources with random emergence at arbitrary locations such as construction, loud music or car alarm.
	However, these temporary non-traffic sources constitute a significant portion of civil noise complaints, rendering a more capable alternative to static noise maps essential for effective noise enforcement \cite{bello2019sonyc}.
	Moreover, relying on noise complaints to mitigate noise violations is a slow process.
	For these reasons, internet of things (IoT) based dynamic noise maps incorporating higher frequency noise level changes are sought for elevated noise enforcement capabilities.

	Dynamic noise maps can be considered as collections of noise maps created at consecutive time steps with short intervals.
	At each time step, either a completely new map or a partial update to the previous map can be created based on the environmental parameters collected from a sensor network.
	No matter what method is used for map generation, existing dynamic mapping research focus on decreasing temporal error of traffic noise and omit the spatiotemporal error caused by non-traffic sources.
	As a result, the noise from non-traffic sources is either filtered out or incorrectly attributed to roads, damaging effectiveness of these maps for noise enforcement.
	
	\IEEEpubidadjcol

	To have a dynamic noise map with a short time step for a low-cost city-wide noise monitoring system, low-power wide-area networks (LPWANs) can be used.
	Although LPWANs provide lower deployment and maintenance costs by eliminating tethered infrastructure via battery-powered wireless nodes, they limit communication data rates significantly.
	For example, LoRaWAN, the most frequently used LPWAN protocol, enables extremely long single-link communication ranges using its LoRa modulation technique in return for extremely low data rates \cite{ikpehai2018low}.
	Combined with hundreds of nodes having to share the same channel using ALOHA-like medium access, the rate and size of messages each node can transmit is significantly constrained to prevent collisions.
	Using only limited data available, estimation of non-traffic source locations on the dynamic map also becomes challenging.
	
	To fill the gap for dynamic maps with non-traffic sources using LPWANs, this study proposes a dynamic mapping system with machine learning (ML) based acoustic localization to overcome the data rate limitation.
	The localization model uses only periodically transmitted 2-byte sound pressure level (SPL) readings and combines them with its prior knowledge of the acoustic behavior of the environment to predict the non-traffic source location.
	The final dynamic map is created as a joint visualization of traffic and non-traffic sources.
	The main contributions of this article can be listed as follows:
	\begin{itemize}
	\item A scalable \lorawan based dynamic noise mapping approach with non-traffic sources is proposed.
	\item An ML based acoustic localization approach using simple SPL indicators to overcome the data rate limitations of LoRaWAN is provided.
	\item A \lorawan acoustic node hardware design targeting low cost noise mapping applications is presented.
	\end{itemize}
	
	The remainder of this article is organized as follows.
	Related work is summarized in Section \ref{sec:related}.
	The details of our proposed system are provided in Section \ref{sec:overview}.
	Performance evaluations based on field-tests are presented in Section \ref{sec:results}.
	Finally, the paper is concluded in Section \ref{sec:conclusion}.

\section{Related Work}
\label{sec:related}
	Existing noise mapping literature can be grouped into simulation and measurement based approaches \cite{liu2020internet}.
	While simulation based approaches use computational methods to estimate noise levels, measurement based approaches use IoT technologies to collect measurements from acoustic sensors to generate a seamless noise level distribution over a wide region.
	Static noise maps are the most prominent examples of simulation based maps and their update interval of several years provide sufficient flexibility to execute time consuming acoustic simulations unsuitable for dynamic maps with frequent updates.
	Using the computational approach, the need to collect scattered SPL measurements are eliminated.
	Studies adopting this approach usually focus on improving the accuracy and computational efficiency of simulations \cite{de2003noise, lesieur2020meta, king2009development}.
	On the other hand, dynamic maps are enabled with measurement based approaches using the spatiotemporal information provided from distributed acoustic sensors.

	A unique IoT-based approach for noise data collection is crowdsensing where microphones and network capabilities of smartphones are used to measure and transmit SPL measurements for map generation \cite{rana2010ear, qin2016noisesense}.
	Although crowdsensing approaches eliminate the usage of a dedicated sensor network, calibration of microphones from different manufacturers and varying availability of up-to-date data may pose significant challenges.
	For example, \cite{picaut2019open} reports almost half of the users removing their application after using it only once.
	Moreover, privacy becomes a significant concern when users' accurate location information is attached to noise measurements \cite{zhang2020tradeoff}.
	
	On the other hand, dedicated acoustic sensor networks with cloud connectivity are used as a more robust IoT backbone for noise measurements, eliminating the shortcomings of crowdsensing in return for increased costs.
	However, existing noise mapping studies focus mostly on demonstrating their mapping techniques using pre-recorded measurements without considering the networking aspect \cite{lesieur2021data, tang2022dynamic, benocci2019dynamic}.
	For example, \cite{wei2016dynamic} demonstrates the integration of SPL readings into static maps using measurements logged at eight locations without any loss of information.
	Similarly, Aumond \etal use SPL values recorded using a mobile measurement device that attaches position information to its logs \cite{aumond2018kriging}.
	Analyzing network conformity is another important aspect for effective evaluation of IoT-based mapping systems to demonstrate their feasibility in real world.
	While many noise monitoring studies prefer spatially sparse SPL values, they do not generate a map with fine-grained representation of noise levels.

	Several noise monitoring studies adopt commercial off-the-shelf IoT-compatible platforms such as Raspberry Pi and BeagleBone as their edge platform \cite{shah2019iot, segura2014low, abeber2018distributed}.
	Although these single-board computers provide elevated capabilities with low cost, their power consumption figures prevent long-term battery-powered operation as reported in \cite{segura2014low} with a daily energy consumption of 43 Wh.
	Hence, \cite{noriega2016application} uses power over ethernet technology to jointly meet network and power requirements of the nodes.
	However, tethered infrastructures are usually not preferred due to their higher long-term costs.
	
	For truly untethered IoT-based noise mapping/monitoring applications, low-power microcontroller based architectures with wireless communications are preferred.
	For this goal, various communication technologies such as WiFi, Zigbee, and \lorawan can be considered \cite{marques2020real, tan2014design, vidana2020low}.
	While WiFi provides the highest data rates, its power consumption is also the highest, preventing long-term battery-powered operation.
	Zigbee on the other hand provides mediocre performance in both data rate and range, while \lorawan or another low-power wide-area network (LPWAN) protocol combines the longest range and low power consumption in return for low data rates \cite{picaut2020low}.
	For example, \cite{domazetovska2020wireless} reports triple the lifetime for their node with \lorawan compared to a similar node using WiFi.
	Thus, \lorawan usually becomes the first choice for non-data-hungry applications.
	Moreover, \cite{gunatilaka2021iot} shows that large datasets can be sent with LPWANs if divided into smaller pieces and sent over longer periods.
	In a different \lorawan based study, LoRa modulation is combined with their unique MAC protocol to enable multi-hop communication unlike the star topology of \lorawan when sending small-sized SPL and class label \cite{yun2022infrastructure}.
	Similarly, \cite{dobrilovic2022urban} uses LoRaWAN to collect SPL measurements to redirect road traffic to decrease noise at noise-sensitive locations such as hospitals.

	No matter which technology is used to transmit noise data, the assumptions on the collected data have a strong impact on the accuracy of generated maps.
	Hence, \cite{benocci2019dynamic} uses a classifier to detect non-traffic noise events and discards audio signals containing these events in SPL calculation to leave traffic noise only for their intended map.
	Although they create event labels in situ, the manual process to remove non-traffic sources prior to SPL calculation prevents it from being realized as a real-time application.
	Conversely, other existing dynamic noise mapping approaches mostly assume contribution from non-traffic sources to noise levels to be negligible, limiting the variety of noise violation types against which dynamic noise maps can be used.

	For a more effective noise enforcement, non-traffic noise sources should also be depicted on the map.
	Such a task is not trivial since the locations of non-traffic sources are arbitrary.
	To overcome this challenge, acoustic localization algorithms can be used to estimate the position of non-traffic sources \cite{cobos2017survey}.
	These algorithms estimate source locations using indicators such as steered response power (SRP), direction of arrival (DoA), energy level, or time difference of arrival (TDoA).
	Due to data rate and energy budget limitations, these algorithms may not be realistic for microcontroller and LPWAN-based IoT systems.

	Among the existing acoustic localization methods, energy based methods offer the lowest computational complexity at the edge nodes while requiring small amount of information transmission (\ie energy level only).
	Least squares and maximum likelihood are among the commonly used methods for energy based acoustic source localization \cite{meng2017energy}.
	However, studies adopting these approaches usually assume free-field propagation \cite{meng2011efficient, sheng2004maximum}.
	In \cite{deng2017energy} and \cite{algobail2019energy}, the localization methods use the energy level ratios of sensor pairs in a non-reverberating environment.
	Similarly, the authors in \cite{correia2021feed} use a feed-forward neural network (FNN) for acoustic localization using energy measurements in free-field environments.
	However, it is shown that omittance of acoustic reflections is known to cause poor performance in urban environments \cite{zhang2024sound}.

	Another IoT-compatible localization indicator is TDoA where cross-correlations of audio signals are used to estimate the time difference of these signals.
	Using TDoA information from multiple sensor nodes, source localization in urban environments becomes possible \cite{hewett2010sound}.
	Despite slightly higher edge computation requirements, TDoA based approach is an important alternative to energy based methods for urban acoustic localization especially when combined with a phase transform (GCC-PHAT) to prioritize phase over magnitude during correlation calculation.
	Instead of using cross-correlation to extract time delay, SRP based localization approaches calculate signal power at candidate source positions and returns the position with the highest SRP as the estimated location \cite{huang2021robust}.
	Since this approach requires cross-correlations to be calculated for all signal pairs, it requires audio signals from all the nodes to be collected at a central location.
	In \cite{huang2021robust} the authors use 2-second audio with 10 kHz sampling rate, thus each node transmits at least 80 KB of data (using 4-byte floating points) for each localization task which would take hours to be transmitted in LPWAN-based systems.

	In \cite{faraji2019sound}, Faraji \etal use DoA for localization, but their DoA calculation method require evaluation of beamformed signal energy in each discrete direction for each pair of 8 microphones whose computational load prevents battery-powered operation beyond 26 hours.
	As an alternative to signal processing based DoA estimation methods, end-to-end ML models eliminate feature extraction step and enable low-computation DoA prediction.
	However, the best offerings of the current literature focus on platforms with significantly higher power consumption than battery-powered nodes used in low-cost IoT-based noise monitoring systems \cite{ko2022real, tan2021extracting}.
	
	\section{System Overview}
	\label{sec:overview}

	In providing a low-cost LPWAN based dynamic noise mapping system, limited data rate of LPWANs poses a big challenge and the system must be designed to overcome this barrier.
	Our proposed system uses ML models to predict the specifications of the non-traffic noise sources on the dynamic map using sparsely transmitted small-sized SPL information from the end nodes.
	Since the acoustic behavior of outdoor environments may vary drastically due to the different sizes and shapes of  physical objects such as buildings, which makes accurate location prediction difficult.
	Therefore, our proposed dynamic noise mapping system divides the application region into 250 m $\times$ 250 m areas and uses area-specific ML models to predict non-traffic noise events as well as their locations and SPLs.
	This way, ML models can learn the acoustic behavior of the specific environment during training, and non-traffic noise sources can be more accurately localized given small amount of SPL data, increasing LPWAN compatibility.
	Such bottom-up approach enables elevated scalability for cities at different sizes.

\begin{figure}
		\includegraphics{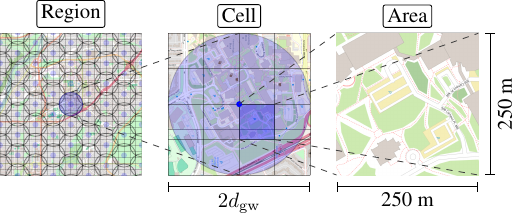}
		\caption{Components of scalable coverage}
		\label{fig:coverage}
	\end{figure}

	In the proposed approach, each area is monitored by nine sensor nodes and the data transmitted by the nodes are collected by multiple gateways.
	Each gateway covers a circular area with a radius of the gateway's communication distance $d_\mathrm{gw}$.
	The frequently used terms of region, cell, and area are visualized in Figure \ref{fig:coverage}.	
	
	For LoRaWAN, although battery-powered nodes eliminate power infrastructure requirement, they may significantly increase maintenance costs if frequent battery replacements are needed.
	To keep power consumption low, the sensor nodes in our proposed system calculate only fast time weighted (L\textsubscript{AF}) and 15-minute long-term equivalent (L\textsubscript{Aeq,15min}) A-weighted SPL values.
	These measurements constitute only 2 bytes in total and are transmitted with only 5 to 15-minute intervals ($\Delta t$).
	As a result, the transmission of these data consumes only between 0.36 to 0.84 bps of physical bitrate for each node when used with $\Delta t = 15$ minutes and DR0 or $\Delta t = 5$ minutes and DR3, respectively.
	
\begin{figure*}
		\includegraphics{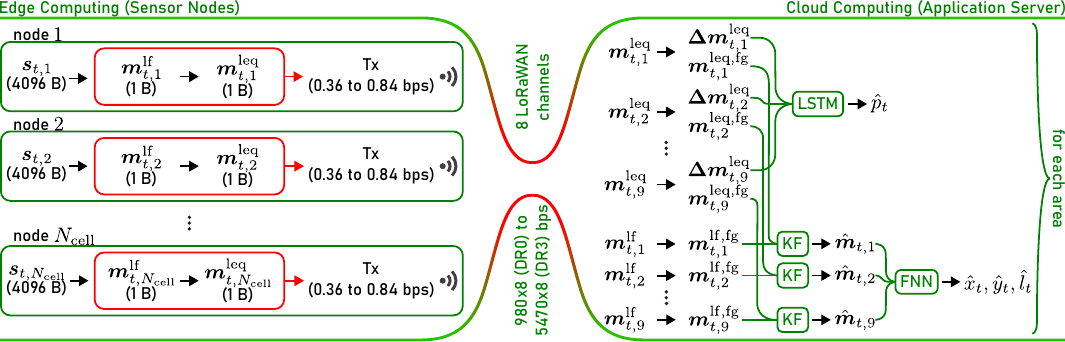}
		\caption{Overview of the proposed dynamic mapping solution to the LoRaWAN data rate bottleneck}
		\label{fig:overview}
	\end{figure*}

	To extract useful information from the scarce sensor data, our system uses two ML models.
	While the first model predicts non-traffic noise events ($\hat{p}$), the other predicts the properties (location and SPL, \ie $\hat{x}, \hat{y}$, and $\hat{l}$) if the first model returns positive.
	This way, computation requirements of the server are decreased by running the secondary model only when a non-traffic noise event exists.
	The predicted non-traffic source is integrated into the map after being simulated as a point source enabling the distinctive output of our system.
	An overview of our proposed approach is shown in Figure \ref{fig:overview}.
	
	When used against noise violations, the time until the dynamic map includes the new noise sources is an important metric.
	In addition to transmission interval, the lack of time synchronization among sensor nodes may contribute to this delay if packet arrivals from all the nodes are waited.
	Moreover, packet losses may cause SPL information temporarily unavailable.
	To overcome these issues, dynamic map update tasks are executed with a 1-minute interval using the latest measurements available to stay functional despite possibly lower output accuracy.
	This approach not only increases system robustness but also eliminates the need for acknowledgements, helping decrease network congestion by reducing the use of valuable downlink windows of LoRaWAN.
	
	It is noted that as a result, each area can have at most one non-traffic noise source at a time.
	Considering the size of an area, the probability of this assumption being violated should be very low and can be omitted.
	Alternatively, the size of the areas can be shrunk until this assumption is fully satisfied.

	We make the sizes of our ML models small so that running the models with a 1-minute interval for small regions can be carried out by regular computers.
	If city-wide coverage with hundreds of areas and 1-minute update interval is desired, a powerful server can be used.
	It is noted that the power required for the proposed system is much less than those mapping systems with complete simulations.

	\subsection{Sensor Node}
\begin{figure}
		\includegraphics{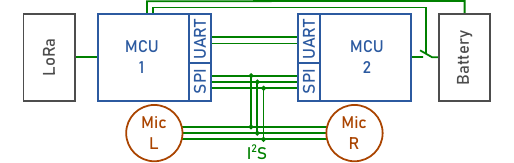}
		\caption{Node hardware}
		\label{fig:node2}
	\end{figure}

\begin{figure}
		\includegraphics{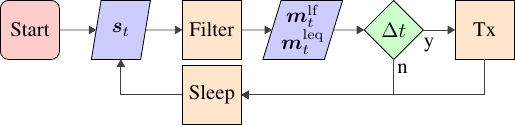}
		\caption{Node firmware state transitions}
		\label{fig:states}
	\end{figure}

	To process acoustic signals and provide extended battery lifetime, we need to develop our own \lorawan acoustic sensor.
	These sensor nodes contain dual microcontrollers, stereo microphones, battery and LoRa transceiver as shown in Figure \ref{fig:node2}.
	All the tasks listed in Figure \ref{fig:states} are performed on the primary microcontroller while the more powerful secondary microcontroller and microphone are reserved for more complicated edge artificial intelligence (AI) functions for our future studies.

	The MEMS microphones located under windscreens at the tip of the extended arms have an overload limit of 120 dB and provide digital audio signal using \isquareds protocol.
	Thanks to this protocol, nodes can easily switch between mono and stereo audio.
	Also, calibration is not mandatory thanks to the $\pm$ 1 dB of sensitivity tolerance \cite{microphone}.
	The nodes first sample audio signal $\bm{s}_t$ at time $t$ with a sampling rate of 31.25 kHz using direct memory access (DMA) for lower power consumption.
	$\bm{s}_t$ contains 1024 samples which takes 4096 bytes in the memory as floating-point values.
	However, the default frequency response of the microphone does not fully lie within the specified tolerances of ANSI specifications for the targeted Type 2 sound level meters and requires an equalizer \cite{ansi}.
	Hence, an equalizing filter with frequency-based gain is applied to move the response at non-conforming frequencies into the specified limits.
	In addition, an A-weighting filter is consecutively applied to better mimic the human perception of audio levels as required by most noise enforcement bylaws.
	These two filters are implemented in second-order recursive structure with minimal memory footprint.
	Following the filtering, L\textsubscript{AF} and L\textsubscript{Aeq} are calculated using:
	\begin{equation}
		\label{eq:laf}
		\bm{m}^{\mathrm{lf}}_{t} = 20\log_{10}
		\left(
		\bm{s}_{\mathrm{RMS},t}
		\middle/
		10^{\displaystyle\left(C_{\mathrm{s}}/20\right)}
		\right)
		+ C_{\mathrm{sr}} + C_{\mathrm{c}
		}
	\end{equation}
	\begin{equation}
		\label{eq:laeq}
		\bm{m}^{\mathrm{leq}}_{t} = 10\log_{10}\left(\displaystyle
		\displaystyle\frac{1}{N_{\mathrm{h}}} \sum_{i=t-N_{\mathrm{h}} + 1}^t 10^{\displaystyle\left(\bm{m}^{\mathrm{lf}}_{i} / 10\right)}\right)
	\end{equation}
	where $C_{\mathrm{s}}$, $C_{\mathrm{sr}}$, and $C_{\mathrm{c}}$ represent microphone sensitivity, reference sensitivity, and calibration constants, respectively.
	In (\ref{eq:laeq}), $T$ represents the preferred time duration for the long-term equivalence and $N_{\mathrm{h}}$ represents the length of $\bm{m}^{\mathrm{lf}}$ storing historical L\textsubscript{AF} values measured within this duration.
	The calculated SPL measurements are within $\pm1$ dB tolerance as tested in L\textsubscript{Aeq,1h}.
	Our system implementation adopts 15-minute L\textsubscript{Aeq} duration as a good trade-off between response latency and accuracy \cite{wei2016dynamic}.
	After the calculation of SPL indicators audio data is deleted, the indicators are transmitted as 1-byte integers if transmission interval is reached.
	Otherwise, the node goes to sleep for 15 seconds as a good tradeoff between measurement accuracy and power consumption \cite{kraemer2019energy}.
	The primary microcontroller can provide power consumption as low as 300 nA in low-voltage retention sleep contributing to daily energy consumption of as low as 50 mWh.
	For detailed consumption analysis readers may refer to \cite{erdem2022energy}.
	
	Using the LoRa transceiver with 125 kHz bandwidth, our nodes can theoretically reach uplink data rates of 980 (with $SF=10$) to 5470 (with $SF=7$) bps on an unoccupied channel.
	Also, our nodes do not utilize automatic data rate (ADR) functionality to preserve valuable downlink channels.

	\subsection{Machine Learning Models}

	\subsubsection{Binary classifier model for non-traffic event detection}
	For event detection $\hat{p}$, a binary classification model using a single long short-term memory (LSTM) layer with 100 units is used to utilize LSTMs' high capability in learning temporal changes.
	The only input feature is the change in the latest discrete L\textsubscript{Aeq} measurement at each node (\ie $\Delta m^{\mathrm{leq}}_{n}$).
	These 9 differences are fed into the network as if they are part of a single feature over discrete time steps instead of individual features to eliminate the need to store historical values and simplify the model.
	The model is trained using an additional dropout layer with 0.5 rate and 1e-4 learning rate.
	
	To save computational resources, the secondary model is executed only when this model predicts a non-traffic noise event.
	Hence, the correct representation of non-traffic source is only possible with high true positive rate of this model.
	
	\subsubsection{Regression model for source property prediction}
	To predict properties (easting ($\hat{x}$), northing ($\hat{y}$), and level ($\hat{l}$)) of the non-traffic source, a regression model using a dual-branch FNN is used.
	Different from the first ML model, the second ML model uses both L\textsubscript{AF} and L\textsubscript{Aeq} foreground measurements fused via Kalman filter (KF) to combine the low latency of L\textsubscript{AF} and low variance of L\textsubscript{Aeq} measurements.
	Here, the term foreground represents the values where background values are subtracted from the measured values (\eg $\bm{m}^{\mathrm{lf,fg}}_{n} = \bm{m}^{\mathrm{lf}}_{n} - \bm{m}^{\mathrm{lf,bg}}_{n}$), with background levels representing the minimum levels observed at each sensor location.

	Since we are using cross-validation with leave-one-out method to evaluate the performance of the proposed system, a secondary set of features is needed to inform the model on the leave-one-out status.
	This way, the ambiguity on whether the measurement being 0 due to the long distance to the source or leave-one-out is eliminated.
	This extra set of features is called mask and is basically an indicator for the presence of the measurement.
	Mask value of zero means the respective measurement is reserved for cross-validation:
	\begin{equation}
		k_{n} = \begin{cases}
		0, & \text{if node $n$ is used for leave-one-out}\\
		1, & \text{otherwise.}
		\end{cases}
	\end{equation}
	Using two different groups of features, we adopt the dual-branch structure so that the network can learn distinct representations for each group before concatenating them.
	For this goal, each feature is connected to a layer of 320 neurons before concatenation.
	The resulting layer is connected to a hidden layer with 100 neurons before the final output layer.
	Between consecutive layers, a dropout rate of 0.2 is used to prevent overfitting.
	The model is trained using a learning rate of 1e-3.
		
	\subsection{Noise Map Generation}
	Our system utilizes individual noise maps for traffic and non-traffic sources before combining them for dynamic map iterations.
	Both traffic and non-traffic maps are generated using NoiseModelling simulation software \cite{bocher2019noisemodelling}.
	While a traffic map uses multiple point sources representing the vehicles on roads, a non-traffic map uses single point source at the predicted location.
	Resulting vector maps are converted to 50 px $\times$ 50 px raster images using the nearest neighbor grid in QGIS software before they are combined \cite{QGIS_software}.
	Since our system divides an entire region into multiple areas, map generation is scripted to adapt different areas based on the coordinates given as parameters.
	
	To embed area-specific acoustic behavior into ML models, each model is trained using noise maps generated for their target area.
	Since collecting high amount of real-world noise violation data is only theoretically possible due to strict legislation around such practice, simulated synthetic data is used in our training.
	Although this approach requires additional computation, it is a one-time preliminary process, and it decreases edge computing complexity enabling localization using only the SPL measurements.

	Since the LSTM model predicts non-traffic noise events, it is trained using a scenario where 500 non-traffic noise sources randomly emerge and disappear within the target area at different times.
	The output maps without reaching a mean value of 35 dB are discarded since they do not possess useful information for training.
	After the resulting map is combined with the base traffic map, $\Delta m^{\mathrm{leq}}_{n}$ values at node locations are extracted with a 1-minute interval.
	
	The second ML model is trained using single-shot noise levels discarding the time domain.
	Random sources in 10,000 trials are used to simulate SPL at sensor nodes.
	The resulting maps are further augmented by adding Gaussian noise ($Z \sim \mathcal{N}(0, 1)$).
	
	Finally, the directivity of the simulated sources is randomized to prevent possible bias.
	The contribution of each frequency bin is randomized while dB magnitude of the directivity ($v$) for each frequency bin at a specific direction ($\theta$) is calculated using:
	\begin{equation}
		\label{eq:directivity}
		v(\theta,i_f) = 40 \left(\frac{\cos(\theta) - 1}{2 i_f + 1}\right)
	\end{equation}
	where $i_f$ $\in$ \{1,2,\ldots,7\} represents frequency bin index for frequencies of \{8000,4000,2000,1000,500,250,125\} Hz, respectively.
	Generated random directivities are fed into NoiseModelling software to be used in simulations.
	Training, validation, and testing splits use 80\%, 20\%, and 20\% of all the simulated map data, respectively.
	
	\subsection{LoRaWAN Communication}
	\label{sec:lora}
	To evaluate system performance, the efficiency of the communication protocol represented as packet success rate (PSR) must be known.
	This study considers two factors affecting PSR: packet collisions due to time overlap and packet losses due to low link quality.
	
	To analyze collision based packet success rate $r_{\mathrm{s,c}}$, packet airtime is needed to analyze time overlaps during channel utilization.
	Packet size in symbols and LoRa symbol rate can be used.
	While LoRa modulation uses a fixed preamble size of 8 symbols, the packet payload size ($n_{\mathrm{pl}}$) varies based on the physical layer payload ($PL$), spreading factor ($SF$), redundancy check ($CRC$), data rate optimization ($DE$), and code rate ($CR$):
	\begin{equation}
		n_{\mathrm{pl}} = 8 + \mathrm{max}\left(0, \left \lceil \frac{2 PL - SF + 7 + 4 CRC}{SF - 2 DE} \right \rceil \left(CR + 4\right) \right)
	\end{equation}

	Assuming that a node uses fixed parameters for packet transmissions, packet airtime ($a_{\mathrm{pkt}}$) is calculated based on preamble and payload sizes ($n_{\mathrm{pa}}=8$ symbols and $n_{\mathrm{pl}}$) as well as symbol rate ($R_\mathrm{s} = BW / 2 ^{SF}$):
	\begin{equation}
		a_{\mathrm{pkt}} = n_{\mathrm{pkt,sym}} / R_{\mathrm{s}}
		= \left(4.25 + n_{\mathrm{pa}} + n_{\mathrm{pl}}\right) / R_{\mathrm{s}}
	\end{equation}
	
	Having packet airtime values, packets from two independent nodes sharing the same channel are assumed to collide if the midpoints of the packets' airtime (\ie $t_x$ and $t_y$) are not as apart as half of their total airtime:
	\begin{equation}
		c(x, y) = \begin{cases}
		1, & \text{if } |t_{x} - t_{y} | < \left(a_{\mathrm{pkt},x} + a_{\mathrm{pkt},y}\right)/2\\
		0, & \text{otherwise.}
		\end{cases}
	\end{equation}
	
	Consequently, the collision based packet success rate in each cell is calculated as the mean collision rate over all the link pairs:
	\begin{equation}
		r_{\mathrm{s,c}} = 1 - \frac{1}{N_\mathrm{cell}\times(N_\mathrm{cell}-1)}\sum_{x \in N_{\mathrm{cell}}}\sum_{\substack{y \in N_\mathrm{cell} \\ y\not = x}} c(x, y)
		\label{eq:psr_c}
	\end{equation}

	The PSR based on link quality $r_{\mathrm{s,q}}$ is calculated considering the unrecoverable received packets due to insufficient SNR or signal strength below the receiver (gateway) sensitivity.
	For this analysis, path loss of link $n$ at distance $d_n$ is calculated based on the mean loss at the reference distance $d_0$, path loss exponent ($\eta$), and shadowing modeled as $X_{\sigma} \sim \mathcal{N}(0, \sigma^2)$:
	\begin{equation}
		L_{\mathrm{p},n}(d_n) = \overline{L_{\mathrm{p}}}(d_0) + 10 \eta \mathrm{log}_{10}\left(d_n / d_0\right) + X_{\sigma}
	\end{equation}

	Using the transmit power ($P_{\mathrm{t}}$), transmitter and receiver antenna gains ($G_{\mathrm{t}}$ and $G_{\mathrm{r}}$), link path loss ($L_{\mathrm{p}}$), bandwidth ($BW$), and noise figure of the receiver hardware ($NF$), received signal power ($P_{\mathrm{r}}$) and its SNR can be calculated using \cite{lin2020experimental}:
	\begin{equation}
		P_{\mathrm{r}} = P_{\mathrm{t}} + G_{\mathrm{t}} + G_{\mathrm{r}} - L_{\mathrm{p}}
	\end{equation}
	\begin{equation}
		SNR = P_{\mathrm{r}} + 174 - 10 \mathrm{log}_{10}\left(BW\right) - NF
	\end{equation}

	The energy per bit to noise power spectral density ratio can be calculated using SNR, spreading factor ($SF$) and $CR$ using:
	\begin{equation}
		\frac{E_{\mathrm{b}}}{N_0} = SNR - 10 \mathrm{log}_{10}\left(\frac{SF \times 4 / (4 + CR)}{2^{SF}}\right)
		\label{eq:snr}
	\end{equation}
	
	Furthermore, link bit error rate is calculated as in \cite{reynders2016range, park2020earn}:
	\begin{equation}
		r_{\mathrm{e,bit}} = Q\left(\frac{\mathrm{log}_{12}(SF)}{\sqrt{2}} \times \frac{E_{\mathrm{b}}}{N_0}\right)
	\end{equation}
	It should be noted that bit error rate is set to 1 for cases where SNR falls below LoRa's demodulation limits or received signal strength is below the modem sensitivity.
	
	To move from bit error rate to packet error rate, packet size in bits must be known.
	This value can be calculated using channel bitrate and packet airtime:
	\begin{equation}
		R_{\mathrm{b}} = \left(SF \times BW\right) / 2^{SF}
	\end{equation}
	\begin{equation}
		n_{\mathrm{pkt,bit}} = \left\lceil a_{\mathrm{pkt}} \times R_{\mathrm{b}}\right\rceil
	\end{equation}

	The conversion from bit error rate to packet success rate for individual link quality is performed with:
	\begin{equation}
		r_{\mathrm{s,q},n} = \left(1 - r_{\mathrm{e,bit}}\right)^{\displaystyle n_{\mathrm{pkt,bit}}}
	\end{equation}
	
	Therefore, the overall link quality based packet success rate is calculated as the average over all the independent links:
	\begin{equation}
		r_{\mathrm{s,q}} = \frac{1}{N_{\mathrm{cell}}}\sum_{n \in N_{\mathrm{cell}}} r_{\mathrm{s,q},n}
		\label{eq:psr_q}
	\end{equation}

	Finally, the PSR of a cell ($r_{\mathrm{s,cell}}$) with $N_{\mathrm{cell}}$ many nodes is calculated as a combination of success rates with time overlap collisions and imperfect link qualities:
	\begin{equation}
		r_{\mathrm{s,cell}} = r_{\mathrm{s,c}} \times r_{\mathrm{s,q}}
		\label{eq:psr_cell}
	\end{equation}
	
	Although we have empirical individual link quality data from the field tests, we are using the above analytical model to define the channel behavior beyond the distances observed in our tests.
	The empirical data is used to verify this model.
	
	As a result, the quality of each link mainly depends on the distance and the $SF$ used.
	This study uses pre-defined assignments ($SF \in \{7,8,9,10\}$) instead of adaptive updates since downlinks diminish the number of possible uplinks significantly.
	For this goal, $SF$ values are assigned based on the link SNR without shadow fading and minimum SNR requirements of LoRa modulation \cite{an1200.22, seller2020predicting}.
	Since varying link quality with shadow fading may cause disconnection, each link uses the next $SF$ (\ie $SF + 1$) whenever possible.
	Although increasing $SF$ by 1 causes longer airtime, it improves link quality beyond the losses caused by time overlap collisions.
	The other parameters used in the channel are as follows: $BW=125$ kHz, $CR=1$, $DE=0$, $P_\mathrm{t}=20$ dBm, $G_{\mathrm{t}}=2.5$ dBi, $G_{\mathrm{r}}=6$ dBi, and $NF=3$ dB.
	For the path loss model, we adopt the parameters from \cite{bor2016lora} (\ie $\eta = 2.08$, $d_0=40$ m, $\overline{L_p}(d_0)=127.41$ dB, and $\sigma=3.57$) since they enable the best fitment to our empirical data.
		
	\section{Experimental Results}
	\label{sec:results}
	
\begin{figure}
		\includegraphics[width=\columnwidth]{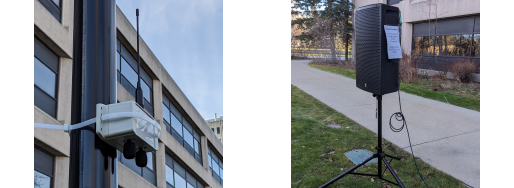}
		\caption{Field test setup}
		\label{fig:setup}
	\end{figure}

	The performance of our proposed system in reducing the error caused by non-traffic source is evaluated in field tests using a speaker shown in Figure \ref{fig:setup} positioned at 6 random locations within the selected application area.
	In each test, the same music tracks were played for 15 minutes with the same order with the sound level of around 100 dBA.
	The reason for such duration is to focus on the entire transitional period of the selected SPL indicator (\ie L\textsubscript{Aeq,15m}).
	Each test was followed by a 15-minute break so that the L\textsubscript{Aeq} measurements can settle back to background levels and allow the equipment to be moved to the next location.

\begin{figure}
		\centering
		\includegraphics{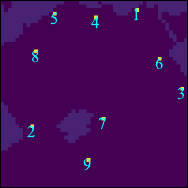}%
		\hspace{0.4625in}
		\includegraphics{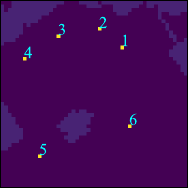}%
		\caption{Sensor (left) and speaker (right) locations}
		\label{fig:locations}
	\end{figure}

	The nodes in the area were deployed as shown in Figure \ref{fig:locations} at 2 m height and at randomly selected existing pole locations that satisfy deployment constraints.
	These constraints force the nodes to be deployed with at least 50 meters distance from each other but within 25 meters of a building facade facing away from the building.
	Although these constraints were not mandatory for this study, they were implemented for a fair performance comparison with our prospective research.
	
	For the field test performance evaluation, the proposed dynamic mapping method was performed in consecutive fashion, where the data for all the field tests were collected and then the mapping was performed on the collected data.
	Although our system can be run in real time, this approach is preferred to augment more field tests with varying transmission intervals and packet success rates to limit the disturbance caused in the tests.
	During field tests, the data was transmitted with a 1-minute interval and then processed for varying transmission intervals and packet losses in 25 trials each.
	In each trial, each node was assigned with a random time offset to better mimic the lack of time synchronization.
	Also, each trial was augmented further by zeroing the measurements from one sensor at a time and setting its mask for cross validation.
	The zeroed values were used as reference values in error calculations.
	Finally, packet loss rates for the respective $\Delta_t$ and $d_\mathrm{gw}$ values were applied.
	As a result, the data collected from 6 field tests was expanded into 16,200 tests.
	
	Table \ref{tab:psr_tot} shows PSRs calculated using (\ref{eq:psr_c}), (\ref{eq:psr_q}), and (\ref{eq:psr_cell}) and the average number of packets reaching the gateway per area per hour.
	Since time-overlap collision probability increases with shorter transmission intervals, $r_\mathrm{s,c}$ decreases with shorter \deltat.
	Moreover, longer transmission distances require higher $SF$, hence longer packet airtime, causing more collisions.
	Thus, the \rsc values decrease as the gateway distance increases.
	On the other hand, link quality is independent from transmission interval.
	Due to higher path losses with longer transmission ranges, \rsq decreases for higher gateway distances despite using higher $SF$ as explained in Section \ref{sec:lora}.
	The best overall cell packet success rate is achieved at 500 m of gateway distance and 15 minutes of transmission interval.
	However, lower transmission intervals enables the highest number of packets reach gateway despite their lower \rscell values, making them more useful for system since they provide more information to the ML models.
	It should be noted that the listed successful packets vary slightly from expected values due to random node deployment and shadowing.

	\begin{table}
		\renewcommand{\arraystretch}{1.3}
		\caption{Performance Results for Lorawan Links}
		\centering
		\begin{tabular}{|c|c|c|c|c|c|}
			\hline
		 	\bfseries $d_{\mathrm{gw}}$ & $\Delta t$ & \bfseries $r_\mathrm{s,c}$ & \bfseries $r_\mathrm{s,q}$ & \bfseries $r_\mathrm{s,cell}$ & \bfseries successful packets \\
			\hline\hline
			\multirow{3}{*}{\rotatebox[origin=c]{90}{500 m}} & 5 min & 0.990 & 0.998 & 0.988 & \bfseries 105 \\
			\cline{2-6}
			& 10 min & 0.997 & 0.998 & 0.995 & 53 \\
			\cline{2-6}
			& 15 min & 0.998 & 0.998 & \bfseries 0.996 & 35 \\
			\cline{2-6}
			\hline
			\multirow{3}{*}{\rotatebox[origin=c]{90}{1000 m}} & 5 min & 0.962 & 0.933 & 0.897 & \bfseries 95 \\
			\cline{2-6}
			& 10 min & 0.981 & 0.933 & 0.915 & 48 \\
			\cline{2-6}
			& 15 min & 0.987 & 0.933 & \bfseries 0.921 & 32 \\
			\cline{2-6}
			\hline
			\multirow{3}{*}{\rotatebox[origin=c]{90}{1500 m}} & 5 min & 0.843 & 0.899 & 0.758 & \bfseries 82 \\
			\cline{2-6}
			& 10 min & 0.917 & 0.899 & 0.824 & 44 \\
			\cline{2-6}
			& 15 min & 0.944 & 0.899 & \bfseries 0.849 & 30 \\
			\cline{2-6}
			\hline
			\multirow{3}{*}{\rotatebox[origin=c]{90}{2000 m}} & 5 min & 0.685 & 0.812 & 0.556 & \bfseries 60 \\
			\cline{2-6}
			& 10 min & 0.827 & 0.812 & 0.671 & 36 \\
			\cline{2-6}
			& 15 min & 0.880 & 0.812 & \bfseries 0.715 & 25 \\
			\cline{2-6}
			\hline
		\end{tabular}
		\label{tab:psr_tot}
	\end{table}
	
	While the performance of our LSTM classification model is evaluated using prediction accuracy, the regression (FNN) model is evaluated based on the mean absolute error (MAE) of its predicted locations.
	In case of false negatives from the first model, our system misses its chance to correct the map error.
	On the other hand, our system may worsen the map error trying to attribute small variances to a noise event by predicting the location of a non-existent source.
	In such cases, the regression model usually predicts only a low source SPL ($\hat{l}$) due to small changes in model inputs, hence the increase in the map error is minimal.
	Nevertheless, prevention of false positives is rendered more important than those of false negatives.
	The performance results of 0.75 precision and 0.57 recall reflect this ambition (\ie higher precision than recall).
	The individual performance results of the ML models are listed in Table \ref{tab:regression} using the metrics of event detection ($\hat{p}$) accuracy, MAE of individual and joint predicted location ($e_{\hat{x}}$, $e_{\hat{y}}$, and $e_{\mathrm{loc}} = \sqrt{{e_{\hat{x}}}^2 + {e_{\hat{y}}}^2}$), as well as MAE of predicted source level ($e_{\hat{l}}$).
	The trends in these metrics are observed to be highly correlated with the number of successful packets presented in Table \ref{tab:psr_tot}.
	More packets reaching the gateway allow more up-to-date information being available for ML model predictions resulting in better location estimation.
	For example, the classifier provides 0.73 event detection accuracy when 105 packets are successfully transmitted, and this value goes down to 0.55 with 35 packets.
	It should be noted that 0.55 accuracy is close to random guess, but such a low accuracy is still useful in decreasing map error thanks to higher precision than recall as well as false positives' destructive impact being negligible due to low source SPL prediction.
	Similarly, location estimation values get better with more packets available.
	The exception of $e_{\hat{l}}$ sometimes getting better with less packets is a result of synthetic data having lower SPL values compared to the real-world data.
	This is also the reason why $e_{\hat{l}}$ is always around 7 dB.
	More successful packets cause the prediction to be better according to the synthetic training data, but this implies a larger discrepancy compared to field-test values.
	Compared with the best acoustic localization algorithms in the literature, 54 to 78 m of location estimation error in a 250 m $\times$ 250 m area may seem large, however the size of the data used to provide these results is extremely small and the data contains information delay up to 15 minutes.
	For the noise mapping application considered here, these localization results are effective in reducing the map error caused by non-traffic sources which is the ultimate goal here.

	\begin{table}
		\renewcommand{\arraystretch}{1.3}
		\caption{Performance Results for ML Models}
		\centering
		\begin{tabular}{|c|c|c|c|c|c|c|}
			\hline
		 	\bfseries $d_{\mathrm{gw}}$ & $\Delta t$ & \bfseries $\hat{p}$ acc. & \bfseries $e_{\hat{x}}$ (m) & \bfseries $e_{\hat{y}}$ (m) & \bfseries $e_{\mathrm{loc}}$ (m) & \bfseries $e_{\hat{l}}$ (dB)\\
			\hline\hline
			\multirow{3}{*}{\rotatebox[origin=c]{90}{500 m}} & 5 min & 0.73 & 32.19 & 40.76 & 54.37 & 7.92\\
			\cline{2-7}
			& 10 min & 0.60 & 37.69 & 41.51 & 59.31 & 7.41\\
			\cline{2-7}
			& 15 min & 0.55 & 44.22 & 48.45 & 70.07 & 6.80\\
			\cline{2-7}
			\hline
			\multirow{3}{*}{\rotatebox[origin=c]{90}{1000 m}} & 5 min & 0.71 & 32.80 & 42.70 & 56.27 & 7.83\\
			\cline{2-7}
			& 10 min & 0.59 & 38.00 & 42.74 & 60.77 & 7.39\\
			\cline{2-7}
			& 15 min & 0.55 & 45.41 & 48.90 & 71.57 & 7.01\\
			\cline{2-7}
			\hline
			\multirow{3}{*}{\rotatebox[origin=c]{90}{1500 m}} & 5 min & 0.68 & 36.39 & 43.82 & 59.95 & 7.55\\
			\cline{2-7}
			& 10 min & 0.59 & 39.61 & 43.03 & 61.96 & 7.43\\
			\cline{2-7}
			& 15 min & 0.56 & 52.50 & 49.55 & 77.88 & 7.39\\
			\cline{2-7}
			\hline
			\multirow{3}{*}{\rotatebox[origin=c]{90}{2000 m}} & 5 min & 0.63 & 42.35 & 47.36 & 68.58 & 7.41\\
			\cline{2-7}
			& 10 min & 0.57 & 47.11 & 39.62 & 67.23 & 7.77\\
			\cline{2-7}
			& 15 min & 0.56 & 49.79 & 47.49 & 75.14 & 7.80\\
			\cline{2-7}
			\hline
		\end{tabular}
		\label{tab:regression}
	\end{table}
	
\begin{figure*}
		\includegraphics{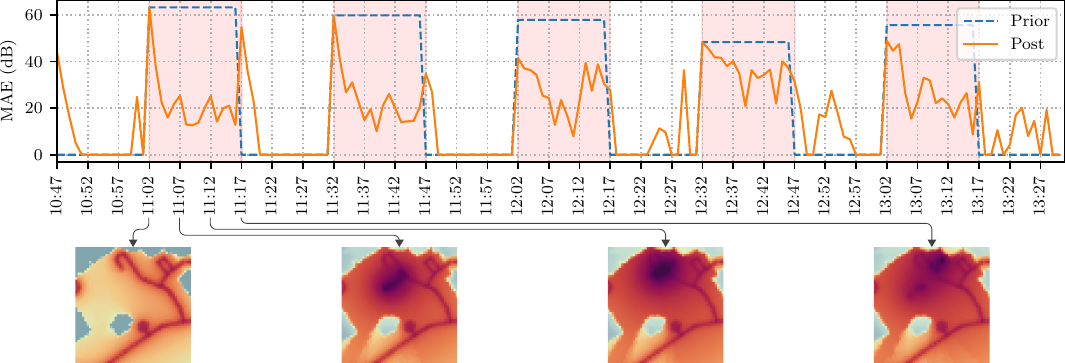}
		\caption{Mean map errors over time for 25 trials (top) and visualized sample maps (bottom) with $\Delta t = 5$ min and $d_\mathrm{gw} = 500$ m}
		\label{fig:error_time}
	\end{figure*}

	Figure \ref{fig:error_time} shows the map error decrease provided by the system as a result of joint utilization of ML models with red background for active test periods in the plot and samples of generated maps below the plot.
	For all the performance evaluations, the first test was clipped to provide equal total duration for different transmission intervals, leaving 5 tests in total.
	Also, cross validation was performed by using the measurements from 1 out of 9 sensors as reference for error calculations and replacing these measurements with zeros for other system tasks in each of 9 cross validation trials.
	For each test, prior error was considered as the maximum minus minimum SPL level throughout the test in dB representing the impact of non-traffic source.
	In addition, post error represents the mean difference in dB between the measured value (1-minute interval data) and the output map value over all the cross validations and random time offset trials at each time step.
	It should be noted that errors can only be measured at node locations since reference measurements are not available outside the node locations.
	As a result, the distance of the source location to the sensor nodes impact how much of the error will be observed.
	In the plot, post error starts with a high value for each test follows a decreasing trend throughout the test thanks to the increasing localization accuracy with more packets arriving and ends with an occasional peak.
	The peaks at the start of each test were caused by the delay until multiple nodes send their measurements with higher SPL values after a non-traffic source emerged.
	During this short transitional period, event detection model predicted false negatives, hence no correction was applied to the map.
	Similarly, the peaks in the error after the source disappears were caused by the delay until nodes' most recently received measurements contained lower values.
	These peaks in prior error last longer when bigger transmission intervals are used.
	These two problematic cases can be observed in sample maps provided at 4 discrete times in Figure \ref{fig:error_time}.
	In the first map only road noise is shown, whereas in reality the test with a non-traffic source started at 11:02.
	Similarly, at 11:17, the source had already stopped making noise, but the presence of the non-traffic source still persisted in the map.
	Also, the occasional peaks between the tests are either caused by false positives in the event detection model or non-controlled noise sources such as birds with loud calls observed during the tests.
	Since our \lorawan acoustic sensors are designed not to store audio recordings for privacy reasons we cannot perform post-analysis to differentiate between these sources and prefer to provide the cumulative error.
	
\begin{figure}
		\includegraphics{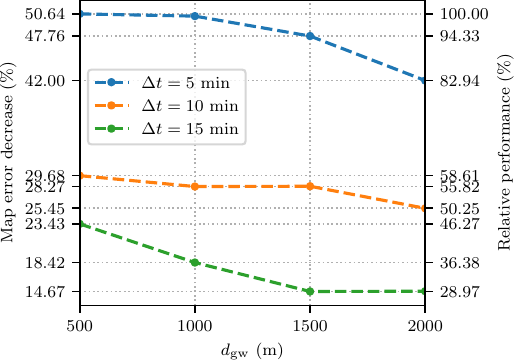}
		\caption{Map error correction results}
		\label{fig:correction}
	\end{figure}

	The mean error correction results throughout all the trials are shown in Figure \ref{fig:correction}.
	The proposed system performs the best with the shortest transmission interval and gateway range.
	Its relative performance drops to 29\% at the longest transmission interval and range.
	Although shorter transmission intervals cause lower PSR, the total number of packets successfully transmitted is still higher and improved system performance is observed as shown in Table \ref{tab:psr_tot}.
	Similarly, a shorter gateway range gives better system performance.
	It should be noted that all the localization and error reduction results are reached using measurements from 8 sensors as a byproduct of cross-validation.

	The proposed system assumes full utilization of 8 out of 72 LoRaWAN uplink channels of US915 specification for city-wide coverage.
	This requirement can be relaxed without compromising the system performance using denser gateways leading to less channel occupation thanks to shorter packet airtimes.
	Although the system already meets the 1\% maximum duty cycle limit asserted in some regions, denser gateways can also be used to adapt the system to the regions with less available channels and/or stricter fair usage limits.

	\section{Conclusion}
	\label{sec:conclusion}
	In this paper, we propose a LoRaWAN based dynamic map for urban noise monitoring.
	We develop a novel LoRaWAN acoustic sensor with edge AI capability.
	Realization of dynamic noise mapping with non-traffic sources is shown to be possible for \lorawan when ML models are utilized to extract useful information from sparsely transmitted data.
	Using experimental data, the proposed system is found to stay operational under high packet collisions when sparse gateway deployment and frequent transmissions are preferred.
	The impact of channel model on this LoRaWAN IoT noise monitoring system is also analyzed.
	
	Our future work is planned to provide improvements to the localization and map error reduction performance using edge ML models while preserving the compatibility with LPWANs.
	
	\bibliographystyle{IEEEtran}
	\bibliography{ref}	

\begin{thebibliography}{10}
\providecommand{\url}[1]{#1}
\csname url@samestyle\endcsname
\providecommand{\newblock}{\relax}
\providecommand{\bibinfo}[2]{#2}
\providecommand{\BIBentrySTDinterwordspacing}{\spaceskip=0pt\relax}
\providecommand{\BIBentryALTinterwordstretchfactor}{4}
\providecommand{\BIBentryALTinterwordspacing}{\spaceskip=\fontdimen2\font plus
\BIBentryALTinterwordstretchfactor\fontdimen3\font minus
  \fontdimen4\font\relax}
\providecommand{\BIBforeignlanguage}[2]{{%
\expandafter\ifx\csname l@#1\endcsname\relax
\typeout{** WARNING: IEEEtran.bst: No hyphenation pattern has been}%
\typeout{** loaded for the language `#1'. Using the pattern for}%
\typeout{** the default language instead.}%
\else
\language=\csname l@#1\endcsname
\fi
#2}}
\providecommand{\BIBdecl}{\relax}
\BIBdecl

\bibitem{passchier2000noise}
W.~Passchier-Vermeer and W.~F. Passchier, ``Noise exposure and public health.''
  \emph{Environmental health perspectives}, vol. 108, no. suppl 1, pp.
  123--131, 2000.

\bibitem{jin2014information}
J.~Jin, J.~Gubbi, S.~Marusic, and M.~Palaniswami, ``An information framework
  for creating a smart city through internet of things,'' \emph{IEEE Internet
  of Things journal}, vol.~1, no.~2, pp. 112--121, 2014.

\bibitem{baclet2022strategic}
S.~Baclet, S.~Venkataraman, R.~Rumpler, R.~Billsj{\"o}, J.~Horvath, and P.~E.
  {\"O}sterlund, ``From strategic noise maps to receiver-centric noise exposure
  sensitivity mapping,'' \emph{Transportation Research Part D: Transport and
  Environment}, vol. 102, p. 103114, 2022.

\bibitem{kephalopoulos2012common}
S.~Kephalopoulos, M.~Paviotti, and F.~Anfosso-L{\'e}d{\'e}e, ``Common noise
  assessment methods in europe ({CNOSSOS-EU}),'' pp. 180--p, 2012.

\bibitem{CRTN}
D.~of~Transport Welsh~Office, \emph{\BIBforeignlanguage{eng}{Calculation of
  road traffic noise}}.\hskip 1em plus 0.5em minus 0.4em\relax London: H.M.S.O.
  London, 1988.

\bibitem{heutschi2018sonroad18}
K.~Heutschi, B.~Locher, and M.~Gerber, ``{sonROAD18}: Swiss implementation of
  the {CNOSSOS-EU} road traffic noise emission model,'' \emph{Acta Acustica
  united with Acustica}, vol. 104, no.~4, pp. 697--706, 2018.

\bibitem{bello2019sonyc}
J.~P. Bello, C.~Silva, O.~Nov, R.~L. Dubois, A.~Arora, J.~Salamon, C.~Mydlarz,
  and H.~Doraiswamy, ``{SONYC}: A system for monitoring, analyzing, and
  mitigating urban noise pollution,'' \emph{Communications of the ACM},
  vol.~62, no.~2, pp. 68--77, 2019.

\bibitem{ikpehai2018low}
A.~Ikpehai, B.~Adebisi, K.~M. Rabie, K.~Anoh, R.~E. Ande, M.~Hammoudeh,
  H.~Gacanin, and U.~M. Mbanaso, ``Low-power wide area network technologies for
  internet-of-things: A comparative review,'' \emph{IEEE Internet of Things
  Journal}, vol.~6, no.~2, pp. 2225--2240, 2018.

\bibitem{liu2020internet}
Y.~Liu, X.~Ma, L.~Shu, Q.~Yang, Y.~Zhang, Z.~Huo, and Z.~Zhou, ``Internet of
  things for noise mapping in smart cities: state of the art and future
  directions,'' \emph{IEEE Network}, vol.~34, no.~4, pp. 112--118, 2020.

\bibitem{de2003noise}
H.~De~Kluijver and J.~Stoter, ``Noise mapping and gis: optimising quality and
  efficiency of noise effect studies,'' \emph{Computers, environment and urban
  systems}, vol.~27, no.~1, pp. 85--102, 2003.

\bibitem{lesieur2020meta}
A.~Lesieur, P.~Aumond, V.~Mallet, and A.~Can, ``Meta-modeling for urban noise
  mapping,'' \emph{The Journal of the Acoustical Society of America}, vol. 148,
  no.~6, pp. 3671--3681, 2020.

\bibitem{king2009development}
E.~King and H.~Rice, ``The development of a practical framework for strategic
  noise mapping,'' \emph{Applied Acoustics}, vol.~70, no.~8, pp. 1116--1127,
  2009.

\bibitem{rana2010ear}
R.~K. Rana, C.~T. Chou, S.~S. Kanhere, N.~Bulusu, and W.~Hu, ``Ear-phone: an
  end-to-end participatory urban noise mapping system,'' in \emph{Proceedings
  of the 9th ACM/IEEE international conference on information processing in
  sensor networks}, 2010, pp. 105--116.

\bibitem{qin2016noisesense}
Z.~Qin and Y.~Zhu, ``Noisesense: A crowd sensing system for urban noise mapping
  service,'' in \emph{2016 IEEE 22nd international conference on parallel and
  distributed systems (ICPADS)}.\hskip 1em plus 0.5em minus 0.4em\relax IEEE,
  2016, pp. 80--87.

\bibitem{picaut2019open}
J.~Picaut, N.~Fortin, E.~Bocher, G.~Petit, P.~Aumond, and G.~Guillaume, ``An
  open-science crowdsourcing approach for producing community noise maps using
  smartphones,'' \emph{Building and Environment}, vol. 148, pp. 20--33, 2019.

\bibitem{zhang2020tradeoff}
Y.~Zhang, M.~Li, D.~Yang, J.~Tang, G.~Xue, and J.~Xu, ``Tradeoff between
  location quality and privacy in crowdsensing: An optimization perspective,''
  \emph{IEEE Internet of Things journal}, vol.~7, no.~4, pp. 3535--3544, 2020.

\bibitem{lesieur2021data}
A.~Lesieur, V.~Mallet, P.~Aumond, and A.~Can, ``Data assimilation for urban
  noise mapping with a meta-model,'' \emph{Applied Acoustics}, vol. 178, p.
  107938, 2021.

\bibitem{tang2022dynamic}
J.-H. Tang, B.-C. Lin, J.-S. Hwang, L.-J. Chen, B.-S. Wu, H.-L. Jian, Y.-T.
  Lee, and T.-C. Chan, ``Dynamic modeling for noise mapping in urban areas,''
  \emph{Environmental Impact Assessment Review}, vol.~97, p. 106864, 2022.

\bibitem{benocci2019dynamic}
R.~Benocci, P.~Bellucci, L.~Peruzzi, A.~Bisceglie, F.~Angelini,
  C.~Confalonieri, and G.~Zambon, ``Dynamic noise mapping in the suburban area
  of {Rome} ({Italy}),'' \emph{Environments}, vol.~6, no.~7, p.~79, 2019.

\bibitem{wei2016dynamic}
W.~Wei, T.~Van~Renterghem, B.~De~Coensel, and D.~Botteldooren, ``Dynamic noise
  mapping: A map-based interpolation between noise measurements with high
  temporal resolution,'' \emph{Applied Acoustics}, vol. 101, pp. 127--140,
  2016.

\bibitem{aumond2018kriging}
P.~Aumond, A.~Can, V.~Mallet, B.~De~Coensel, C.~Ribeiro, D.~Botteldooren, and
  C.~Lavandier, ``Kriging-based spatial interpolation from measurements for
  sound level mapping in urban areas,'' \emph{The journal of the acoustical
  society of America}, vol. 143, no.~5, pp. 2847--2857, 2018.

\bibitem{shah2019iot}
S.~K. Shah, Z.~Tariq, and Y.~Lee, ``Iot based urban noise monitoring in deep
  learning using historical reports,'' in \emph{2019 IEEE International
  Conference on Big Data (Big Data)}.\hskip 1em plus 0.5em minus 0.4em\relax
  IEEE, 2019, pp. 4179--4184.

\bibitem{segura2014low}
J.~Segura-Garcia, S.~Felici-Castell, J.~J. Perez-Solano, M.~Cobos, and J.~M.
  Navarro, ``Low-cost alternatives for urban noise nuisance monitoring using
  wireless sensor networks,'' \emph{IEEE Sensors Journal}, vol.~15, no.~2, pp.
  836--844, 2014.

\bibitem{abeber2018distributed}
J.~AbeBer, M.~Gotze, S.~Kuhnlenz, R.~Grafe, C.~Kuhn, T.~ClauB, and
  H.~Lukashevich, ``A distributed sensor network for monitoring noise level and
  noise sources in urban environments,'' in \emph{2018 IEEE 6th International
  Conference on Future Internet of Things and Cloud (FiCloud)}.\hskip 1em plus
  0.5em minus 0.4em\relax IEEE, 2018, pp. 318--324.

\bibitem{noriega2016application}
J.~E. Noriega-Linares and J.~M. Navarro~Ruiz, ``On the application of the
  raspberry pi as an advanced acoustic sensor network for noise monitoring,''
  \emph{Electronics}, vol.~5, no.~4, p.~74, 2016.

\bibitem{marques2020real}
G.~Marques and R.~Pitarma, ``A real-time noise monitoring system based on
  internet of things for enhanced acoustic comfort and occupational health,''
  \emph{IEEE Access}, vol.~8, pp. 139\,741--139\,755, 2020.

\bibitem{tan2014design}
W.~M. Tan and S.~A. Jarvis, ``On the design of an energy-harvesting
  noise-sensing wsn mote,'' \emph{EURASIP Journal on Wireless Communications
  and Networking}, vol. 2014, pp. 1--18, 2014.

\bibitem{vidana2020low}
E.~Vida{\~n}a-Vila, J.~Navarro, C.~Borda-Fortuny, D.~Stowell, and R.~M.
  Alsina-Pag{\`e}s, ``Low-cost distributed acoustic sensor network for
  real-time urban sound monitoring,'' \emph{Electronics}, vol.~9, no.~12, p.
  2119, 2020.

\bibitem{picaut2020low}
J.~Picaut, A.~Can, N.~Fortin, J.~Ardouin, and M.~Lagrange, ``Low-cost sensors
  for urban noise monitoring networks—a literature review,'' \emph{Sensors},
  vol.~20, no.~8, p. 2256, 2020.

\bibitem{domazetovska2020wireless}
S.~Domazetovska, M.~Anachkova, V.~Gavriloski, and Z.~Petreski, ``Wireless
  acoustic low-cost sensor network for urban noise monitoring,'' in \emph{Forum
  Acusticum}, 2020, pp. 677--682.

\bibitem{gunatilaka2021iot}
D.~Gunatilaka, ``An iot-enabled acoustic sensing platform for noise pollution
  monitoring,'' in \emph{2021 IEEE 12th Annual Ubiquitous Computing,
  Electronics \& Mobile Communication Conference (UEMCON)}.\hskip 1em plus
  0.5em minus 0.4em\relax IEEE, 2021, pp. 0383--0389.

\bibitem{yun2022infrastructure}
J.~Yun, S.~Srivastava, D.~Roy, N.~Stohs, C.~Mydlarz, M.~Salman, B.~Steers,
  J.~P. Bello, and A.~Arora, ``Infrastructure-free, deep learned urban noise
  monitoring at\~{} 100mw,'' in \emph{2022 ACM/IEEE 13th International
  Conference on Cyber-Physical Systems (ICCPS)}.\hskip 1em plus 0.5em minus
  0.4em\relax IEEE, 2022, pp. 56--67.

\bibitem{dobrilovic2022urban}
D.~Dobrilovi{\'c}, V.~Brtka, G.~Jotanovi{\'c}, {\v{Z}}.~Stojanov,
  G.~Jau{\v{s}}evac, and M.~Mali{\'c}, ``The urban traffic noise monitoring
  system based on lorawan technology,'' \emph{Wireless Networks}, vol.~28,
  no.~1, pp. 441--458, 2022.

\bibitem{cobos2017survey}
M.~Cobos, F.~Antonacci, A.~Alexandridis, A.~Mouchtaris, B.~Lee \emph{et~al.},
  ``A survey of sound source localization methods in wireless acoustic sensor
  networks,'' \emph{Wireless Communications and Mobile Computing}, vol. 2017,
  2017.

\bibitem{meng2017energy}
W.~Meng and W.~Xiao, ``Energy-based acoustic source localization methods: a
  survey,'' \emph{Sensors}, vol.~17, no.~2, p. 376, 2017.

\bibitem{meng2011efficient}
W.~Meng, W.~Xiao, and L.~Xie, ``An efficient em algorithm for energy-based
  multisource localization in wireless sensor networks,'' \emph{IEEE
  Transactions on Instrumentation and Measurement}, vol.~60, no.~3, pp.
  1017--1027, 2011.

\bibitem{sheng2004maximum}
X.~Sheng and Y.-H. Hu, ``Maximum likelihood multiple-source localization using
  acoustic energy measurements with wireless sensor networks,'' \emph{IEEE
  transactions on signal processing}, vol.~53, no.~1, pp. 44--53, 2004.

\bibitem{deng2017energy}
F.~Deng, S.~Guan, X.~Yue, X.~Gu, J.~Chen, J.~Lv, and J.~Li, ``Energy-based
  sound source localization with low power consumption in wireless sensor
  networks,'' \emph{IEEE Transactions on Industrial Electronics}, vol.~64,
  no.~6, pp. 4894--4902, 2017.

\bibitem{algobail2019energy}
A.~Algobail, A.~Soudani, and S.~Alahmadi, ``Energy-efficient scheme for target
  recognition and localization in wireless acoustic sensor networks,''
  \emph{International Journal of Distributed Sensor Networks}, vol.~15, no.~11,
  p. 1550147719891406, 2019.

\bibitem{correia2021feed}
S.~D. Correia, S.~Tomic, and M.~Beko, ``A feed-forward neural network approach
  for energy-based acoustic source localization,'' \emph{Journal of Sensor and
  Actuator Networks}, vol.~10, no.~2, p.~29, 2021.

\bibitem{zhang2024sound}
D.~Zhang, J.~Chen, J.~Bai, and M.~Wang, ``Sound event localization and
  classification using wasn in outdoor environment,'' \emph{arXiv preprint
  arXiv:2403.20130}, 2024.

\bibitem{hewett2010sound}
D.~Hewett, ``Sound propagation in an urban environment,'' Ph.D. dissertation,
  Oxford University, UK, 2010.

\bibitem{huang2021robust}
Y.~Huang, J.~Tong, X.~Hu, and M.~Bao, ``A robust steered response power
  localization method for wireless acoustic sensor networks in an outdoor
  environment,'' \emph{Sensors}, vol.~21, no.~5, p. 1591, 2021.

\bibitem{faraji2019sound}
M.~M. Faraji, S.~B. Shouraki, E.~Iranmehr, and B.~Linares-Barranco, ``Sound
  source localization in wide-range outdoor environment using distributed
  sensor network,'' \emph{IEEE Sensors Journal}, vol.~20, no.~4, pp.
  2234--2246, 2019.

\bibitem{ko2022real}
J.~Ko, H.~Kim, and J.~Kim, ``Real-time sound source localization for low-power
  {IoT} devices based on multi-stream {CNN},'' \emph{Sensors}, vol.~22, no.~12,
  p. 4650, 2022.

\bibitem{tan2021extracting}
E.-L. Tan, F.~A. Karnapi, L.~J. Ng, K.~Ooi, and W.-S. Gan, ``Extracting urban
  sound information for residential areas in smart cities using an end-to-end
  iot system,'' \emph{IEEE Internet of Things Journal}, vol.~8, no.~18, pp.
  14\,308--14\,321, 2021.

\bibitem{microphone}
\BIBentryALTinterwordspacing
``\BIBforeignlanguage{en-US}{{ICS}-43434 {\textbar} {TDK}}.'' [Online].
  Available: \url{https://invensense.tdk.com/products/ics-43434/}
\BIBentrySTDinterwordspacing

\bibitem{ansi}
{American National Standards Institute}, ``{ANSI} {S1}.4: {Specifications} for
  {Sound} {Level} {Meters}.''

\bibitem{kraemer2019energy}
F.~A. Kraemer, F.~Alawad, and I.~M.~V. Bosch, ``Energy-accuracy tradeoff for
  efficient noise monitoring and prediction in working environments,'' in
  \emph{Proceedings of the 9th International Conference on the Internet of
  Things}, 2019, pp. 1--8.

\bibitem{erdem2022energy}
H.~E. Erdem, H.~Leung, and N.~Xie, ``Energy neutral urban noise monitoring and
  classification with {LoRaWAN} based {IoT},'' in \emph{2022 IEEE
  Sensors}.\hskip 1em plus 0.5em minus 0.4em\relax IEEE, 2022, pp. 1--4.

\bibitem{bocher2019noisemodelling}
E.~Bocher, G.~Guillaume, J.~Picaut, G.~Petit, and N.~Fortin, ``Noisemodelling:
  An open source {GIS} based tool to produce environmental noise maps,''
  \emph{Isprs international journal of geo-information}, vol.~8, no.~3, p. 130,
  2019.

\bibitem{QGIS_software}
\BIBentryALTinterwordspacing
{QGIS Development Team}, ``{QGIS} geographic information system,'' QGIS
  Association, 2023. [Online]. Available: \url{https://www.qgis.org}
\BIBentrySTDinterwordspacing

\bibitem{lin2020experimental}
K.~Lin and T.~Hao, ``Experimental link quality analysis for lora-based wireless
  underground sensor networks,'' \emph{IEEE Internet of Things Journal},
  vol.~8, no.~8, pp. 6565--6577, 2020.

\bibitem{reynders2016range}
B.~Reynders, W.~Meert, and S.~Pollin, ``Range and coexistence analysis of long
  range unlicensed communication,'' in \emph{2016 23rd International Conference
  on Telecommunications (ICT)}.\hskip 1em plus 0.5em minus 0.4em\relax IEEE,
  2016, pp. 1--6.

\bibitem{park2020earn}
J.~Park, K.~Park, H.~Bae, and C.-K. Kim, ``Earn: Enhanced adr with coding rate
  adaptation in lorawan,'' \emph{IEEE Internet of Things Journal}, vol.~7,
  no.~12, pp. 11\,873--11\,883, 2020.

\bibitem{an1200.22}
{Semtech Corporation}, ``{AN1200}.22 {LoRa} {Modulation} {Basics},'' May 2015.

\bibitem{seller2020predicting}
O.~Seller, ``Predicting {LoRaWAN} {Capacity},'' May 2020.

\bibitem{bor2016lora}
M.~C. Bor, U.~Roedig, T.~Voigt, and J.~M. Alonso, ``Do lora low-power wide-area
  networks scale?'' in \emph{Proceedings of the 19th ACM international
  conference on modeling, analysis and simulation of wireless and mobile
  systems}, 2016, pp. 59--67.

\end{thebibliography}

	\vskip -2\baselineskip plus -1fil
	
\end{document}